\documentclass[11pt]{article}
\usepackage[margin=1in]{geometry}
\usepackage[T1]{fontenc}
\usepackage[utf8]{inputenc}
\usepackage{booktabs}
\usepackage{longtable}
\usepackage{array}
\usepackage{graphicx}
\usepackage{fancyhdr}
\usepackage{hyperref}
\usepackage{xcolor}

\hypersetup{
  colorlinks=true,
  linkcolor=blue,
  urlcolor=blue,
  pdftitle={A Behavioral State Vocabulary in Sony ERS-111 R-CODE},
  pdfauthor={Christopher A. Tucker},
  pdfsubject={Embodied behavior vocabularies, Sony ERS-111 R-CODE, and robotic control architecture},
  pdfkeywords={AIBO, ERS-111, R-CODE, behavior diagrams, embodied AI, robotics}
}

\title{A Behavioral State Vocabulary in Sony ERS-111 R-CODE}
\author{Christopher A. Tucker\\
\small \texttt{cartheur@pm.me}\\
\small \href{https://orcid.org/0000-0002-0172-7258}{ORCID: 0000-0002-0172-7258}}
\date{}

\begin{document}
\maketitle

\begin{abstract}
This paper presents a corpus-level analysis of generated behavior
diagrams derived from Sony's \texttt{R-CODE} sample distribution for the
\texttt{ERS-111} \texttt{AIBO}. Rather than reading each script in
isolation, the study compares named states across the corpus to
identify the recurring control vocabulary that structures the sample
set. The resulting aggregate shows that many superficially different
routines are built from a compact embodied grammar centered on
initialization, sensing, iterative action, synchronization, and
recovery. It further shows that this vocabulary supports a graded scale
of rising behavioral complexity, from capability activation and startup
regularization to monitored locomotion, environmental decision loops,
and fuller mode-based control. In addition to historical analysis, the
paper argues that this form of state-based abstraction is useful as an
intermediate representation for constructing new encapsulated behavior
routines, especially on constrained native robotic systems where
deterministic control, direct hardware access, and modular behavioral
composition remain important.
\end{abstract}

\section*{Introduction}

This paper examines the generated \texttt{ERS-111} \texttt{R-CODE}
behavior diagrams as a single behavioral corpus rather than as an
assortment of independent scripts. Its objective is to identify the
common control vocabulary that structures the sample set,
distinguishing recurring foundational states from those that appear
only within more specialized variants.

This framing is useful in part because many contemporary robotics
systems still exhibit a persistent imbalance between deliberation and
embodiment. They may perform well in perception benchmarks, scripted
manipulation pipelines, or high-level planning tasks, yet remain
comparatively weak in the continuous background competencies that allow
an animal or human to remain viable in the world while doing something
else. As a heuristic analogy, one may think here of a layered
\texttt{System 1} and \texttt{System 2} organization: fast,
low-latency, embodied monitoring and response on one side, and slower,
more reflective or task-selective decision processes on the other. In
the psychological literature, these terms generally distinguish rapid,
automatic, and largely background cognition from slower, more
deliberative reasoning \cite{kahneman2011thinking}. Many current robot architectures are still
stronger on the deliberative
side than on the deep integration of these everyday embodied loops.
The value of the \texttt{R-CODE} corpus, therefore, is not that it
solves that modern problem in full, but that it gives a compact
historical example of behavior being organized around persistent bodily
monitoring, recovery, local branching, and return-to-loop coordination
rather than around isolated commands alone \cite{kawaharazuka2024realworld,xu2024survey}.

The paper makes four main contributions. First, it provides a corpus
aggregate over the currently generated \texttt{ERS-111} diagrams and
identifies the most common state titles and their frequencies. Second,
it uses that aggregate to argue that the corpus is organized around a
compact, reactive, embodied state vocabulary rather than around
unrelated one-off routines. Third, it shows that this vocabulary has a
graded internal structure: across examples running from capability
activation, to startup regularization, to monitored locomotion, to
local interaction branching, to obstacle-dependent decision loops, and
finally to task-specific mode control, the corpus provides a visible
scale of rising behavioral complexity. Fourth, it treats that
abstraction not only
as an analytic device for legacy code, but also as a practical design
representation for constructing new behavior routines on native robotic
systems \cite{arkin1998behavior,brooks1991intelligence}.

\section*{Method and Corpus}

These totals provide the first numerical outline of the corpus. Their
value is not solely descriptive. They help determine whether the sample
set is broad but structurally diffuse, or whether many distinct scripts
are constructed from a comparatively compact set of recurring state
forms. In this respect, the totals establish a baseline for assessing
how much behavioral variation is produced from how little underlying
control vocabulary.

The following corpus totals are included as a compact numerical
reference point for the discussion that follows. They identify the size
of the diagram set, the total number of state occurrences observed
across it, and the number of distinct state titles from which that
behavioral variation is constructed.

The method used here is deliberately simple and explicit. Behavior
diagrams were generated from the \texttt{ERS-111} sample corpus, state
titles were normalized where placeholder labels still obscured their
functional role, and the resulting state names were then counted
across the generated set. The aim is not to claim that the titles
alone exhaust the semantics of the original code, but to show that the
state layer is already rich enough to expose a stable behavioral
grammar at corpus scale. Once that aggregate was established,
representative diagrams were then compared as a graded sequence in
order to test whether the same shared vocabulary also supports a
visible progression of increasing control complexity across different
behavior families.

\begin{itemize}
\item diagrams counted: \texttt{54}
\item total state instances: \texttt{292}
\item unique state titles: \texttt{47}
\end{itemize}

The state-title aggregate makes it possible to treat the Sony
\texttt{ERS-111} \texttt{R-CODE} samples as a behavioral corpus rather than
as an assortment of independent scripts. The most immediate result is
the repeated reuse of a small control vocabulary. The most common
titles, especially \texttt{Sense / Decide}, \texttt{Action Loop},
\texttt{Boot / Safe Pose}, \texttt{Synchronize}, \texttt{Boot},
\texttt{Sense Fall State}, and \texttt{Recover}, indicate that the
dominant form in the corpus is not an extended linear routine. Instead,
it is a recurrent embodied loop in which the robot establishes a viable
posture, samples some portion of its body or local environment,
selects an action, allows that action to settle, and then returns to
monitoring.

This finding is significant because it indicates that the surface
diversity of the scripts exceeds their structural diversity. A football
behavior, an obstacle routine, a contact-response behavior, and a
locomotion loop may differ in immediate purpose, yet the aggregate
shows that they are often composed from the same recurring internal
elements. The corpus is therefore not simply a set of demonstrations of
what AIBO can do; it is also evidence of how Sony's \texttt{R-CODE}
sample distribution organized behavior. The repetition of these core
titles suggests a practical design grammar in which behavior is
assembled through recombination of a small number of stable
state-types.

The highest-frequency title, \texttt{Sense / Decide}, is particularly
important. Its prominence suggests that much of the corpus is organized
around conditional interpretation rather than predetermined motion
playback alone. The scripts repeatedly evaluate the present situation
and determine which state should follow. This places the sample set
closer to reactive robotics than to simple animation. When considered
alongside the strong presence of \texttt{Action Loop} and
\texttt{Synchronize}, it points to a common sequence: evaluate, act,
wait for bodily completion, and reevaluate. This is a compact yet
robust pattern for embodied behavior \cite{arkin1998behavior,brooks1991intelligence}.

The frequent appearance of \texttt{Boot} and \texttt{Boot / Safe Pose}
is likewise meaningful. These states show that initialization is not
treated as an incidental preface but as part of the logic of behavior
itself. The robot is repeatedly brought into a known physical and
control condition before more specific behavior begins. Even these
relatively early or simple scripts therefore assume that behavior
depends on a stable bodily baseline. In robotics terms, the robot's
initial pose and internal setup form part of the control architecture
rather than a mere setup convenience.

The recurring appearance of \texttt{Sense Fall State} and
\texttt{Recover} is among the most revealing findings. It indicates
that bodily failure is not peripheral within the corpus. The robot's
viability as a moving body is part of the logic that organizes
behavior. The code does not merely issue movements and assume continued
stability; it repeatedly checks whether posture has degraded and routes
control through recovery when necessary. This reflects a strongly
embodied orientation in the sample set. The robot is managing itself as
a physical agent in the world, not merely executing isolated commands.

The lower-frequency state titles are equally informative because they
show where specialization enters the corpus. Titles such as
\texttt{Left Kick}, \texttt{Right Kick}, \texttt{Start Ball Tracking},
and \texttt{It approaches a ball by the angle of the head} belong to
narrower families such as \texttt{Football} and related variants. Their
lower counts indicate that these states do not belong to the
corpus-wide backbone. Instead, they function as family-specific
elaborations layered onto the more common loop of sensing, action,
synchronization, and recovery. This distinction helps separate general
architecture from task-specific extension.

The same holds for titles such as \texttt{Happy Hug Response},
\texttt{Idle Thought}, \texttt{Baby Sway Forward},
\texttt{Select First Average}, and \texttt{Reset Sample Count}. These
are analytically useful because they show how the shared control
vocabulary is directed toward different ends: social contact,
expressive motion, temporary local computation, and directional
averaging, among others. Their rarity, however, indicates that they are
local elaborations rather than dominant organizational principles. In
aggregate, the sample set is therefore less a collection of unrelated
unique behaviors than a set of behaviors constructed from a relatively
compact repertoire of reusable state-forms.

The elimination of placeholder titles was analytically important.
Names such as \texttt{State 200}, \texttt{State 1011},
\texttt{State 1110}, and \texttt{State 1200} obscured the functional
meaning the aggregate was meant to recover. These were temporary labels
introduced during extraction when numeric state identifiers from the
source scripts had not yet been resolved into functional names.
Renaming them as
\texttt{Quit Behavior}, \texttt{Reset Sample Count},
\texttt{Select First Average}, and \texttt{Reset Scan Counter} made
those states legible as behavioral units rather than residual labels
from the original scripts. Once the placeholders were removed, the
count no longer mixed structural meaning with accidental naming
residue. The aggregate therefore became not merely a quantitative list,
but a more reliable semantic map of the corpus.

Taken together, the aggregate supports a clear interpretation: the
\texttt{ERS-111} \texttt{R-CODE} samples are organized around a compact
embodied state vocabulary that is repeatedly recombined across
different behavior families. The counts show that the common
foundation is reactive and bodily grounded, while the rarer states
indicate where the families diverge into specialized task or expressive
behavior. This provides a clearer basis for discussing the material in
both historical and technical terms. Historically, it suggests that
Sony's samples already embodied a recognizable control grammar.
Technically, it indicates that many apparently distinct behaviors can
be understood as variations on a shared loop of setup, sensing,
decision, action, stabilization, and recovery.

The aggregate also carries broader significance for robotics. State
titles and transitions are not useful only for analyzing legacy AIBO
scripts; they express a more general principle of robot control.
Advanced systems likewise distinguish among modes such as searching,
tracking, avoiding, recovering, and interacting. What changes over time
is the richness of those states, the sophistication of their triggers,
and the layering of control above and below them. In this respect, the
\texttt{ERS-111} sample corpus offers a small but concrete example of
how complex robotic behavior can emerge through repeated reuse of a
compact state-transition architecture \cite{arkin1998behavior,brooks1991intelligence}.

The tables below make that distinction easier to see directly. The first
table captures the high-frequency state titles that form the recurring
structural backbone of the corpus. The second table gathers the
lower-frequency titles that appear more locally, often within narrower
behavior families or single specialized variants.

\section*{Most Common State Titles}

\begin{center}
\begin{tabular}{>{\ttfamily}r l}
\toprule
Count & State Title \\
\midrule
56 & Sense / Decide \\
51 & Action Loop \\
36 & Boot / Safe Pose \\
32 & Synchronize \\
18 & Boot \\
14 & Sense Fall State \\
14 & Recover \\
13 & Head Scan Position \\
8 & Repeat Forward Walk \\
4 & Left Kick \\
4 & Right Kick \\
3 & It approaches a ball by the angle of the head \\
\bottomrule
\end{tabular}
\end{center}

These more common titles provide the clearest expression of the shared
behavioral grammar that runs through the sample set. They show where
Sony's examples repeatedly return to the same embodied control
concerns: establishing posture, assessing conditions, selecting
actions, waiting for movement to settle, and recovering when the body
is no longer in a viable state.

\section*{Secondary State Titles}

\begin{center}
\begin{longtable}{>{\ttfamily}r p{4.3in}}
\toprule
Count & State Title \\
\midrule
\endhead
2 & Quit Behavior \\
2 & Start Ball Tracking \\
2 & Mode \\
2 & Slow Forward Walk \\
1 & Main \\
1 & Increment Lost Counter \\
1 & Reset Lost Counter \\
1 & Set Search Mode \\
1 & Turn Left \\
1 & Shift Left \\
1 & Advance Left \\
1 & Search Turn \\
1 & Shift Right \\
1 & Advance Right \\
1 & Forward Walk \\
1 & Happy Hug Response \\
1 & It sleeps until it is set up \\
1 & Idle Thought \\
1 & Baby Sway Forward \\
1 & Baby Sway Backward \\
1 & When there was an obstacle in front \\
1 & Reset Sample Count \\
1 & When Distance became less than 300 \\
1 & At the time of loop\_cnt=1 \\
1 & At the time of loop\_cnt=2 \\
1 & Wait loop end \\
1 & Select First Average \\
1 & Reset Scan Counter \\
1 & Start \\
1 & Load Stored Poses \\
1 & Save Current Poses \\
1 & Resume Listener \\
1 & MainLoop \\
1 & Clear Light Playback \\
1 & Power Down \\
\bottomrule
\end{longtable}
\end{center}

Taken together, the secondary titles show where the common architecture
branches into more specific local roles. They capture narrower aspects
of the corpus: expressive reactions, temporary counters, intermediate
scan steps, branch-specific motor decisions, and other states that are
important within individual families without dominating the corpus as a
whole.

\section*{Representative Behavior Diagrams}

This section presents the paper's second line of evidence in visual
form. Alongside the corpus-level aggregate, it offers a graded
behavioral ladder that makes one of the paper's stronger claims
directly visible: the shared control vocabulary of the \texttt{R-CODE}
corpus is not only recurrent, but scalable. It supports a progression
from minimal activation and startup organization, through monitored
movement, contact-conditioned response, and obstacle-dependent decision
loops, to fuller task-specific mode control. Read in that order, the
figures do not merely illustrate isolated scripts. They demonstrate
that the corpus contains a usable scale of rising behavioral
complexity, and therefore a more precise basis for comparing
sophistication across routines.

\begin{figure}[h]
\centering
\includegraphics[width=0.86\textwidth]{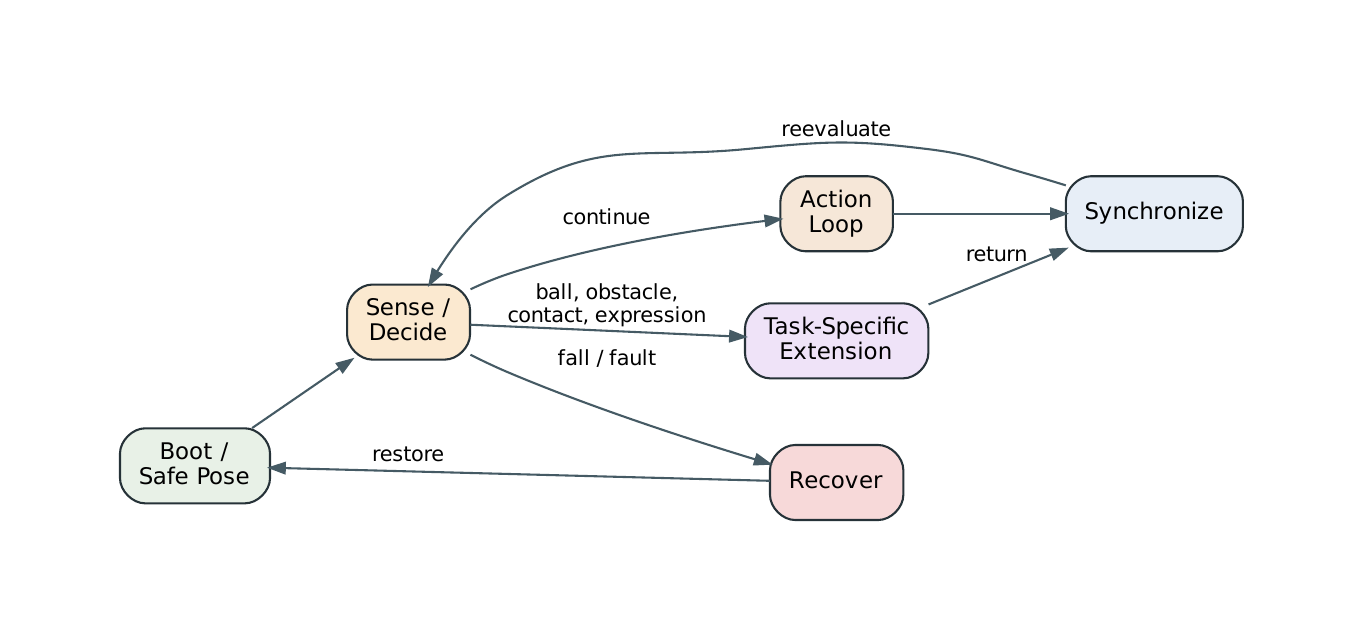}
\caption{The embodied behavior aggregate provides a corpus-level view of
the recurring control vocabulary. It makes visible the repeated
dependence on shared states such as sensing, looping, synchronization,
and recovery across otherwise distinct behaviors.}
\end{figure}

\begin{figure}[h]
\centering
\begin{tabular}{ccc}
\includegraphics[width=0.20\textwidth]{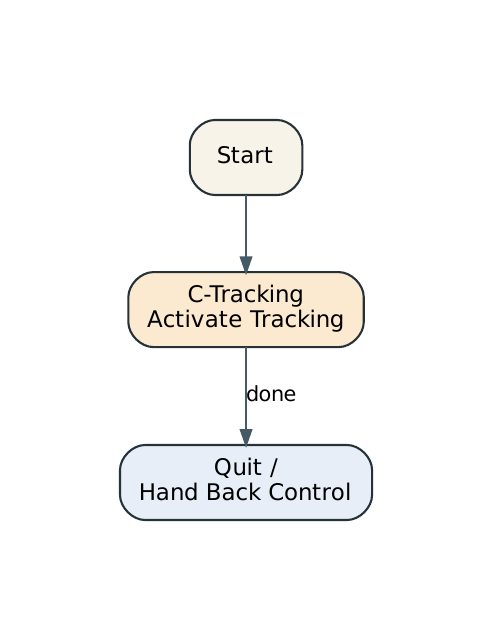} &
\includegraphics[width=0.21\textwidth]{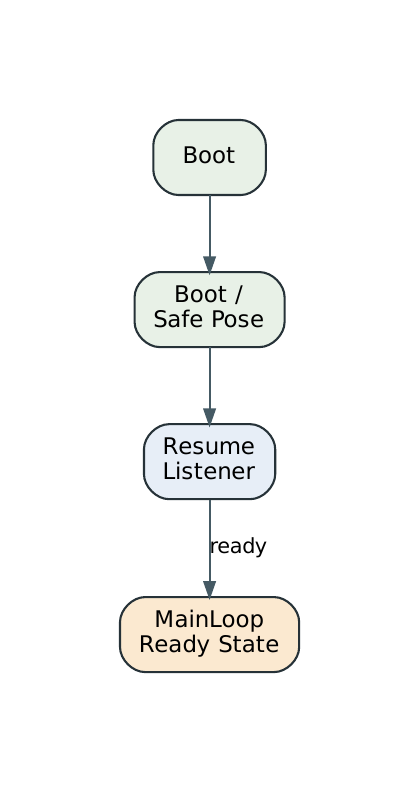} &
\includegraphics[width=0.24\textwidth]{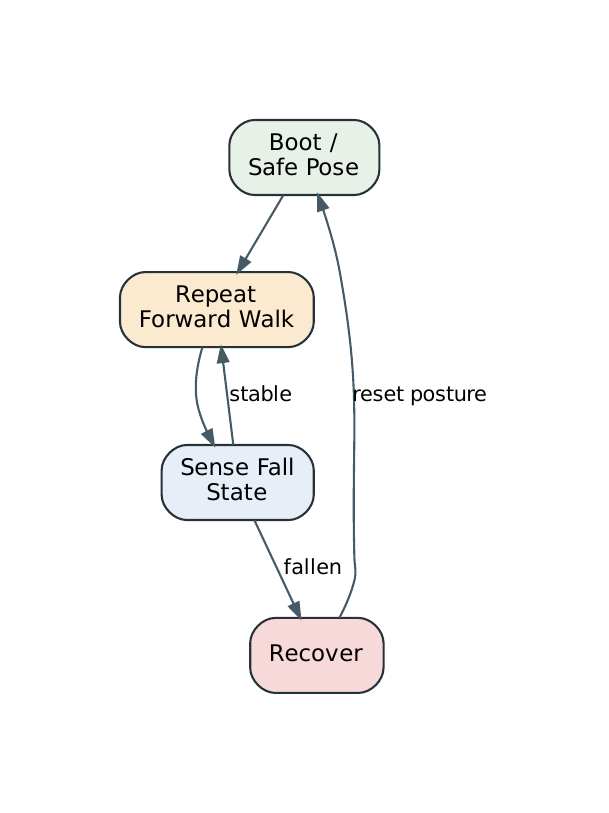} \\
\texttt{C-Tracking} &
\texttt{Boot / Safe Pose} &
\texttt{Move} \\
\includegraphics[width=0.24\textwidth]{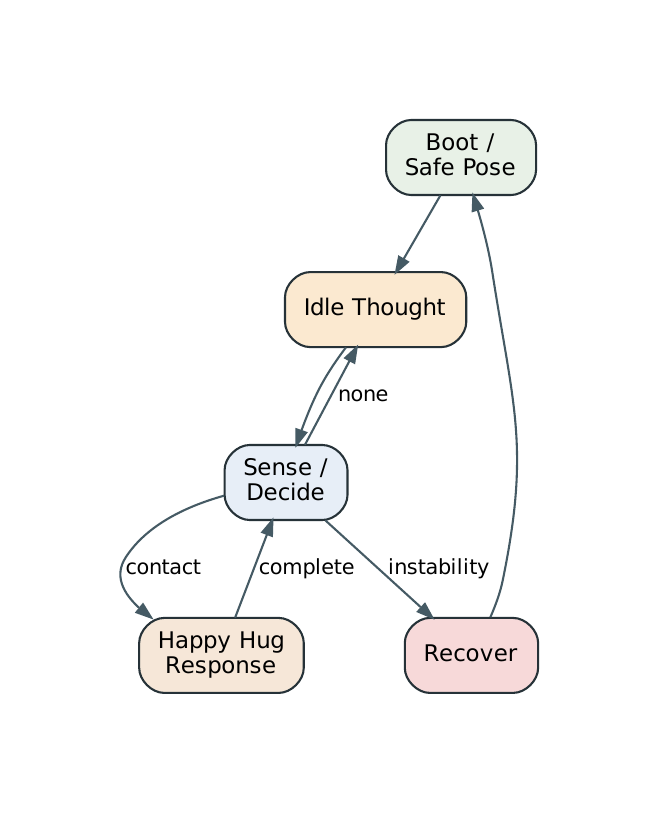} &
\includegraphics[width=0.28\textwidth]{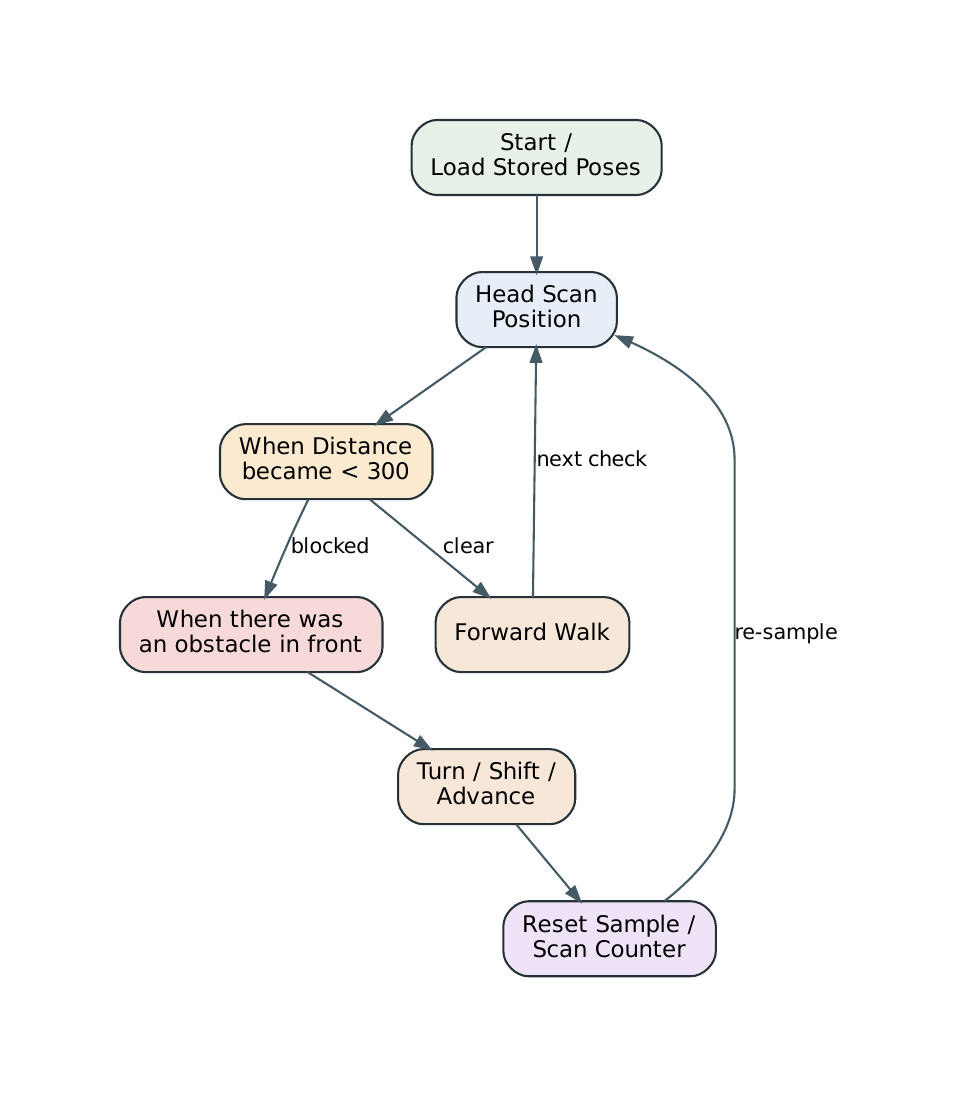} &
\includegraphics[width=0.22\textwidth]{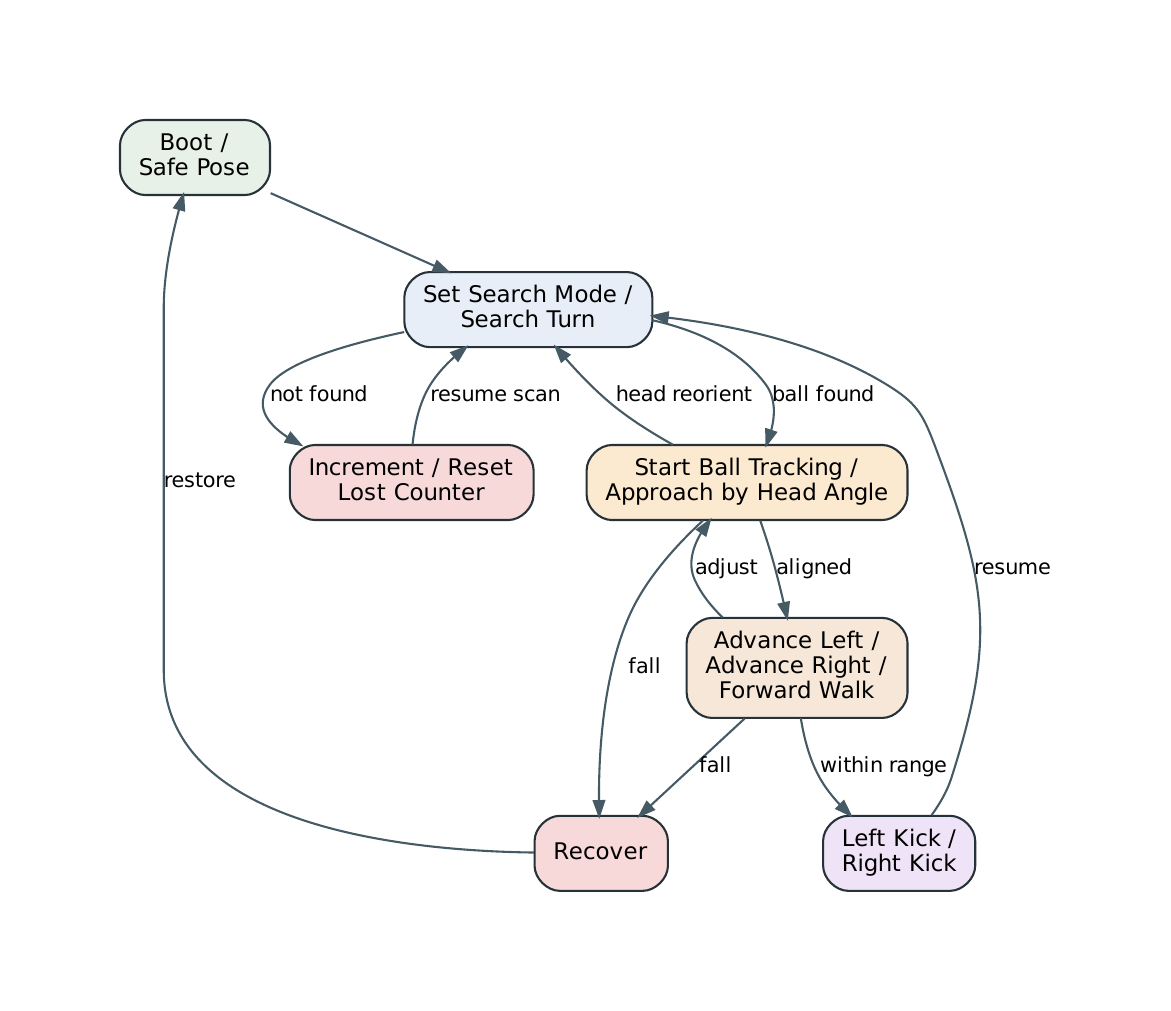} \\
\texttt{Contact Response} &
\texttt{Maze} &
\texttt{Football}
\end{tabular}
\caption{A compact comparison ladder from minimal activation to explicit
mode control. Read left to right, top to bottom, the reductions show a
graded increase in behavioral organization: capability activation,
startup regularization, monitored locomotion, local interaction
branching, obstacle-dependent decision loops, and finally multi-state
task control.}
\end{figure}

Read analytically, each step in the ladder adds a distinct control
commitment. \texttt{C-Tracking} stands closest to capability
activation: it enters a single role and then yields control back with
little visible internal differentiation. \texttt{Boot / Safe Pose}
already deepens the picture by showing that initialization is itself a
behavioral unit: before task execution, the robot is brought into a
known bodily and control condition. \texttt{Move} adds persistent
embodied monitoring, since forward locomotion is no longer treated as a
one-shot act but as a loop checked against fall state and routed
through recovery when posture degrades. \texttt{Contact Response}
introduces local conditional branching around interaction, showing that
the same shared backbone can be redirected toward expressive or social
response rather than locomotion alone.

\texttt{Maze} marks the important intermediate threshold at which the
behavior becomes visibly decision-structured. Here the robot does not
merely continue, stop, or recover. It repeatedly samples the local
environment, compares distance conditions, redirects action when an
obstacle appears, and then re-enters a sampling cycle. That pattern is
more than reactive repetition; it is a compact decision loop organized
around environmental comparison and controlled reorientation.
\texttt{Football} extends that same logic into fuller task-specific
mode control. Search, ball tracking, approach, kick selection, loss
handling, and recovery appear as linked but differentiated
sub-behaviors. The analytical value of the sequence is therefore not
just that later diagrams contain more states, but that one can watch
the growth from activation, to initialization, to monitored action, to
conditional interaction, to environmental decision, and finally to
persistent multi-state embodied organization.

\section*{The Purpose of Aggregation}

This aggregate is intended to show which control structures are broadly
shared across the \texttt{ERS-111} \texttt{R-CODE} set and which belong
only to narrower behavior families. The counts are analytically useful
not because they rank scripts by popularity, but because they reveal
the shared behavioral grammar of the corpus: the points at which the
code repeatedly returns to sensing, looping, synchronization, and
recovery before adding more specialized states for pursuit, contact
response, or expressive variation.

This analysis is relevant beyond \texttt{AIBO}, because the same
pattern appears in many embodied robotic systems: a stable loop of
sensing, deciding, acting, and rechecking the body and local
environment before continuing. Viewed in this way, the
\texttt{ERS-111} samples are not only historically interesting
artifacts. They also provide an early small-scale example of how
robotics often develops in practice, by building richer task-specific
states on top of a compact base of posture control, environmental
monitoring, transition handling, and recovery \cite{kawaharazuka2024realworld,xu2024survey}.

The principal result is not simply that one state title appears more
often than another. The stronger result is that the diagram set
repeatedly converges on the same architectural pattern:

\begin{center}
\texttt{initialize -> sense -> decide -> act -> synchronize -> repeat}
\end{center}

This indicates that the preserved \texttt{ERS-111} samples already
express a recognizable embodied control style. Even when the surface
behaviors appear different, the underlying workflow is usually a
compact reactive cycle rather than a one-shot animation or an extended
scripted narrative. The graded comparison ladder strengthens this claim
by showing that the same recurring backbone does not merely repeat. It
also admits ordered elaboration, so that increasingly complex routines
can be compared as staged extensions of a common control base.

A second important result is that bodily maintenance appears within the
recurring structure itself. The regular presence of \texttt{Sense Fall
State} and \texttt{Recover} shows that the robot's posture and
stability are treated as part of behavior control rather than as
external exceptions. In analytical terms, the corpus concerns not only
the issuance of motion, but the maintenance of viable motion in the
body.

A third result is that the corpus provides an explicit scale of rising
behavioral sophistication rather than a flat inventory of scripts.
\texttt{C-Tracking}, \texttt{Boot / Safe Pose}, \texttt{Move},
\texttt{Contact Response}, \texttt{Maze}, and \texttt{Football} do not
only differ in theme. They expose successive increases in control
organization: activation, initialization, monitored continuity,
interaction branching, environmental decision structure, and persistent
mode control. This graded structure makes the sample set more useful
analytically because it gives the reader a principled way to compare
behavior families by depth of organization rather than by task label
alone.

A fourth result is that the more specialized behaviors remain legible
because they are layered on top of that common base. Pursuit and kick
states in the \texttt{Football} family, for example, do not replace the
core reactive loop; they extend it. Complexity in the sample set tends
to accumulate by adding narrower decision and action states to an
already stable embodied architecture.

Taken together, these findings clarify the central purpose of this
paper. The \texttt{R-CODE} sample set can be read as an early
behavioral design vocabulary in which a limited number of state-types
are repeatedly recombined to produce locomotion, orientation, contact
response, recovery, and more specialized goal-directed action. The
importance of the aggregate, therefore, lies not simply in documenting
individual scripts, but in revealing the shared organizational logic
that makes those scripts analytically comparable. What may initially
appear to be a heterogeneous collection of sample behaviors instead
emerges as a coherent control corpus whose recurring state structures
offer a foundation for historical interpretation, comparative
analysis, and broader discussion of embodied robot behavior. Just as
important, the graded ladder shows that this shared logic has internal
depth: the corpus supports a visible progression from simple activation
to richer state-governed behavioral organization. That progression is a
core contribution of the paper because it turns the corpus into a scale
for reasoning about increasing control complexity rather than a mere
catalog of examples.

This approach is also constructive rather than merely descriptive.
Once the common structure of a behavior corpus has been made explicit,
those states can be treated as reusable design units for new robotic
systems. Instead of writing a new routine as a monolithic block of
control code, one can begin from a small library of encapsulated
behavioral forms: initialize posture, sense local conditions, branch by
event, execute a bounded action, verify bodily state, and return to a
monitoring loop. The diagrams therefore function not only as
retrospective analysis, but as a practical intermediate representation
for synthesis. They help make clear which elements should remain stable
across platforms and which may be replaced to accommodate different
sensors, actuators, timing models, or task domains \cite{arkin1998behavior}.

That design value becomes especially important in native-hardware
robotics developed in a polyForth style. In such environments,
behavior is often assembled from compact words, explicit stacks, direct
device access, and tightly bounded control loops rather than from large
framework abstractions. A state-based representation is well matched to
that discipline because each state can be implemented as a small,
testable routine with clear entry conditions, side effects, and exit
transitions. The transition structure then serves as the behavioral
scheduler. This makes it easier to preserve determinism, reason about
timing, and isolate hardware-specific code inside a limited number of
sensing and actuation words while leaving the higher-level behavioral
organization intact.

For newer classes of robot systems, this means the method supports both
portability and invention. A locomotion platform, a manipulator, or a
socially interactive robot may differ greatly in embodiment, but each
still benefits from an architecture that separates posture management,
environmental interpretation, action execution, exception handling, and
return-to-loop coordination. By treating those elements as explicit
behavioral compartments, developers can introduce characteristic new
routines without losing architectural clarity. In practical terms, one
can retain the same behavioral skeleton while substituting new state
contents for vision, force sensing, gait control, manipulation,
dialogue, or safety recovery. The result is a disciplined path for
building novel robot behavior on constrained native systems without
collapsing all control logic into a single opaque procedure.

\section*{Conclusion}

States and transitions are significant because they provide a robot
with a means of organizing time, context, and response rather than
reacting through a flat sequence of commands. A state indicates the
mode in which the robot is currently operating, and a transition
indicates the condition sufficient to move it into another mode. This
structure becomes a foundation for more advanced robotics because
sensing, motion, recovery, and task logic can remain coordinated rather
than collapsing into an undifferentiated control stream.

The present paper has argued that the preserved \texttt{ERS-111}
\texttt{R-CODE} corpus already makes this principle visible in a
practical historical setting. Its aggregate shows a recurring embodied
control vocabulary, and its graded comparison ladder shows that this
vocabulary scales from capability activation and startup
regularization, through monitored movement and environmental decision
loops, to more specialized mode-governed forms. On that basis, the
corpus provides more than historical documentation. It offers a compact
behavioral design language and a visible scale of control complexity
that can support both interpretation of early robot software and
construction of new native robotic routines. The strongest conclusion
is therefore not only that the \texttt{ERS-111} sample set reuses a
common behavioral grammar, but that it makes increasing behavioral
sophistication analytically legible as a graded progression of control
organization. That claim gives the corpus its broader value: it can be
used not just to catalog legacy routines, but to compare, interpret,
and eventually reconstruct increasingly rich embodied behaviors on a
shared architectural basis.

\vspace{-0.5\baselineskip}
\bibliographystyle{unsrt}
\bibliography{references}

\end{document}